\documentclass{article}
\usepackage{spconf,amsmath,epsfig}

\usepackage{threeparttable}
\usepackage{multirow}
\usepackage{booktabs}
\usepackage{amssymb}

\begin{document}\sloppy

\def\x{{\mathbf x}}
\def\L{{\cal L}}

\title{AMS-SFE: Towards an Alignment of Manifold Structures via Semantic Feature Expansion for Zero-shot Learning}
%
\name{Jingcai Guo, Song Guo \thanks{This work is supported by National Natural Science Foundation of China (61872310) and INTPART BDEM project.}}
\address{Department of Computing, The Hong Kong Polytechnic University, Kowloon, Hong Kong.\\
cscjguo@comp.polyu.edu.hk, song.guo@polyu.edu.hk}

\maketitle

\begin{abstract}
Zero-shot learning (ZSL) aims at recognizing unseen classes with knowledge transferred from seen classes. This is typically achieved by exploiting a semantic feature space (FS) shared by both seen and unseen classes, i.e., attributes or word vectors, as the bridge. However, due to the mutually disjoint of training (seen) and testing (unseen) data, existing ZSL methods easily and commonly suffer from the domain shift problem. To address this issue, we propose a novel model called AMS-SFE. It considers the Alignment of Manifold Structures by Semantic Feature Expansion. Specifically, we build up an autoencoder based model to expand the semantic features and joint with an alignment to an embedded manifold extracted from the visual FS of data. It is the first attempt to align these two FSs by way of expanding semantic features. Extensive experiments show the remarkable performance improvement of our model compared with other existing methods.
\end{abstract}
\section{Introduction and Motivation}
Zero-shot learning (ZSL), which aims to imitate human ability in recognizing unseen classes, has received increasing attention in the most recent years \cite{frome2013devise,shigeto2015ridge,bucher2016improving,zhang2016zero,kodirov2017semantic,zhu2018generative,guo2019adaptive}. ZSL takes utilization of labeled seen class examples and certain knowledge that can be transferred and shared between seen and unseen classes. This knowledge, e.g., attributes, exist in a high dimensional vector space called semantic feature space (FS). The attributes are meaningful high-level information about examples, such as their shapes, colors, components, textures, etc. Intuitively, the cat is more closely related to the tiger than to the snake. In the semantic FS, this intuition also exists. The similar classes have similar patterns in the semantic FS, and this particular pattern is called the prototype. Each class is embedded to the semantic FS and endowed with a prototype. As a common practice in ZSL, an unseen class example is first projected from the visual FS to the semantic FS by a projection trained on seen classes. Then with such obtained semantic features, we search the most closely related prototype whose corresponding class is set to this example. Specifically, this relatedness can be measured by the similarity or distance between the semantic features and prototypes.
\begin{figure}[t]
    \centerline{\includegraphics[width=0.37\textwidth]{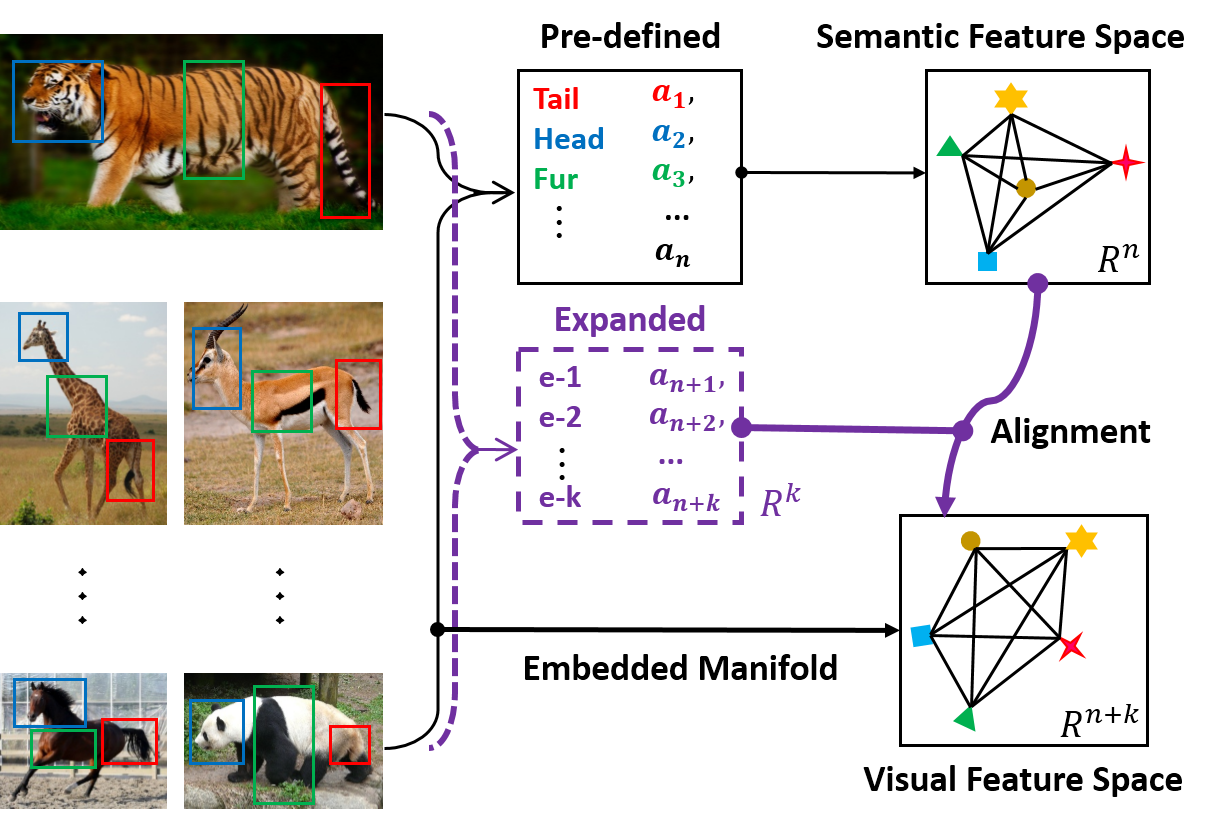}}
     \caption{The framework of proposed AMS-SFE}
    \label{fig1}
    \vspace{-0.2cm} 
\end{figure}

However, due to the absence of unseen classes when training this projection, the domain shift problem \cite{fu2015transductive} easily happens. Moreover, the visual FS and the semantic FS are mutually independent and different. Therefore, it is challenging to obtain a well-matched projection between the visual and the semantic FSs. In this paper, to address the above issues, we propose a novel model to align the manifold structures from the semantic FS to the visual FS (Fig.1). Specifically, we train an autoencoder based model which takes the visual features as inputs and generates $k$-dimensional auxiliary features for each prototype in the semantic FS, except for the pre-defined $n$-dimensional features. Meanwhile, we make use of these auxiliary semantic features and combine with the pre-defined ones to discover the better adaptability for semantic FS. This is mainly achieved by aligning the manifold structures from the combined semantic FS ($\mathbb{R}^{n+k}$) to an embedded $(n+k)$-dimensional manifold extracted from the original visual FS. The expansion and alignment phases are conducted simultaneously by joint supervision from both the reconstruction and the adaptation terms within the autoencoder based model. 

Our model results in two benefits. (1) Since the prototypes are typically pre-defined by experts, or by algorithms from the external resource \cite{lampert2014attribute,fu2015transductive}, they may have some limitations to adapt to more substantial and new scenarios with increasing classes in the real-world situation. By properly expanding some auxiliary semantic features, we can enhance the representation capability of semantic FS. (2) More importantly, by utilizing these expanded auxiliary features, we can implicitly align the manifold structures from the semantic to visual FSs. Because of the two benefits, our model can obtain a more robust projection that greatly mitigates the domain shift problem, and generalize better to unseen classes.



\section{Related Work}

The domain shift problem is firstly identified and studied by Fu et al. \cite{fu2015transductive}, which refers to the phenomenon that when projecting unseen class examples from the visual FS to the semantic FS, the obtained results easily shift away from the real ones (prototypes). This is essentially caused by the nature of ZSL that the training (seen) and testing (unseen) classes are mutually disjoint. Due to the absence of unseen classes during training, it is challenging to obtain a well-matched projection or to do domain adaptation with unseen classes.

Recently, inductive learning based methods \cite{fu2015zero,changpinyo2016synthesized} and transductive learning based methods \cite{fu2015transductive} have been investigated. The former enforces additional constraints from the training data, while the latter assumes that the unseen class examples (unlabelled) are available at once during training. Generally speaking, the performance of the latter is better than the former because of the utilization of extra information from unseen classes. However, the transductive learning is not fully complied with the zero-shot setting that no example from unseen classes is available during training. The manifold learning is based on the idea that there exists a lower-dimensional manifold embedded in a high dimensional space. Recently, the semantic manifold distance \cite{fu2015zero} is introduced to redefine the distance metric in the semantic FS using a novel absorbing Markov chain process. MFMR \cite{xu2017matrix} leverages the sophisticated technique of matrix tri-factorization with manifold regularizers to enhance the projection between visual and semantic spaces. With the popularity of generative adversarial networks (GANs), some related ZSL methods have also been proposed. GANZrl \cite{AAAI1816805} applies GANs to synthesize examples with specified semantics to cover a higher diversity of seen classes. Instead, GAZSL \cite{zhu2018generative} leverages GANs to imagine unseen classes from text descriptions. 

Despite the progress made, the domain shift problem is still an open issue. In our model, we consider expanding some auxiliary semantic features to implicitly align the semantic and visual FSs. Similar to GANs, the expansion phase in our model is also a generative task but focuses on the semantic feature level, and our autoencoder based model is lighter and easier to implementation yet effective. Moreover, we strictly comply with the zero-shot setting that the training is solely based on the seen class examples.

\section{Proposed Method}

\subsection{Problem Definition}

We start by formalizing the zero-shot learning task and then introduce our proposed method and formulation. Given a set of labeled seen class examples $\mathcal{D}=\left \{ x_{i}, y_{i} \right \}_{i=1}^{l}$, where $x_{i}\in\mathbb{R}^{d}$ is a seen class example as visual features with class label $y_{i} \in C=\left \{ c_{1}, c_{2}, \cdots , c_{m}\right \}$. The goal is to build a model for a set of unseen classes ${C}' = \left \{ {c}'_{1}, {c}'_{2}, \cdots , {c}'_{v}\right \}$ ($C \bigcap {C}' = \phi$) which have no labeled examples during training. In the testing phase, given a test example ${x}'\in\mathbb{R}^{d}$, the model predicts its class label $c({x}')\in{C}'$. To this end, some bridging information (i.e., the semantic features), denote as $S^{p} = \left ( a_{1}, a_{2}, \cdots , a_{n} \right )\in \mathbb{R}^{n}$, is needed in ZSL as common knowledge in the semantic FS, where each dimension $a_{i}$ is one specific feature or property. Therefore, the seen class examples can be further specified as $\mathcal{D}=\left \{ x_{i}, y_{i}, S^{p}_{i} \right \}_{i=1}^{l}$. Each seen class $c_{i}$ is endowed with a semantic prototype $P^{p}_{c_{i}}\in \mathbb{R}^{n}$, and each unseen class ${c_{i}}'$ is also endowed with a semantic prototype ${P^{p}_{{c_{i}}'}}'\in \mathbb{R}^{n}$. Thus for each seen class example we have $S^{p}_{i} \in P^{p}=\left \{P^{p}_{c_{1}}, P^{p}_{c_{2}}, \cdots , P^{p}_{c_{m}}  \right \}$, while for testing unseen classes, we need to predict their semantic features ${S^{p}}' \in \mathbb{R}^{n}$ and set their class labels by searching the most closely related prototypes within ${P^{p}}'=\left \{{P^{p}_{{c_{1}}'}}', {P^{p}_{{c_{2}}'}}', \cdots , {P^{p}_{{c_{v}}'}}'  \right \}$.

\subsection{Method and Formulation}

\subsubsection{Semantinc Feature Expansion}

To align the manifold structures from semantic to visual FS, the first step of our model is to expand the semantic features. Specifically, we keep the pre-defined semantic features $S^{p}=(a_{1}, a_{2}, \cdots , a_{n})\in \mathbb{R}^{n}$ fixed and expand extra $k$-dimensional auxiliary semantic features $S^e = \left (a_{n+1}, a_{n+2}, \cdots , a_{n+k}\right )\in \mathbb{R}^{k}$. We build an autoencoder based network to extract these features. Each seen class example $x_{i} \in \mathbb{R}^{d}$ is encoded to a latent feature vector $z_{i} \in \mathbb{R}^{k} (k\ll d)$ by $Encoder(x_{i})$ in the auxiliary FS. Then followed by a decoder, the network reconstructs this example as $\hat{x_{i}} \in \mathbb{R}^{d}$ by $Decoder(z_{i})$. In this step, the reconstruction loss can be described as:
\begin{equation}
\begin{small}
\mathcal{L}_r = \sum_{i=1}^{l}\left \| x_{i} - \hat{x_{i}} \right \|_{2}^{2}\,.
\end{small}
\end{equation}
We minimize it to guarantee the learned latent vector $z_{i}$ retains the most potent information of the input $x_{i}$.

\subsubsection{Embedded Manifold Extraction}

Before exploiting these auxiliary semantic features, we extract a lower-dimensional embedded manifold ($\mathbb{R}^{n+k}$) of the visual FS ($\mathbb{R}^{d}$, $n+k\ll d$) to utilize the structure information. We first find and define the center of each seen class in the visual FS as $x^{c} = \left \{x^{c_{i}} \right \}_{i=1}^{m}$, where $\left \{c_{i} \right \}_{i=1}^{m}$ are $m$ class labels and $x^{c_{i}}$ is the center (i.e., the mean value) of all examples belonging to class $c_{i}$. Then we compose a matrix $\mathbf{D}=\left [ d_{ij} \right ]\in \mathbb{R}^{m\times m}$ that records the distance of each center-pair in the visual FS, where $d_{i,j} = \left \| x^{c_{i}} - x^{c_{j}} \right \|$. Then we search for a lower-dimensional embedded manifold ($\mathbb{R}^{n+k}$) that can be modeled by $(n+k)$-dimensional embedded features. We denote the embedded representation matrix of centers as $\mathbf{O}=\left [ o_{i} \right ]\in \mathbb{R}^{(n+k) \times m}$, and expect $\mathbf{O}$ retains the geometrical and distribution constraints of the visual FS. A natural idea is that the distance matrix $\mathbf{D}$ also restrains the embedded representation matrix $\mathbf{O}$, that the distance of each center-pair $\left \| o^{c_{i}} - o^{c_{j}} \right \|$ in the corresponding FS ($\mathbb{R}^{(n+k)}$) also holds.

To this end, we denote the inner product of $\mathbf{O}$ as $\mathbf{B} = \mathbf{O}^\top \mathbf{O}\in \mathbb{R}^{m \times m}$, so that $b_{ij} = o_i^\top o_j$ and we can obtain:
\begin{equation}
\begin{small}
\begin{aligned}
d_{ij}^2 &= \left \| o_i \right \|^2 + \left \| o_j \right \|^2-2o_i^\top o_j = b_{ii} + b_{jj} - 2b_{ij}\,,
\end{aligned}
\end{small}
\end{equation}
We set $\sum_{i=1}^{m}o_{i}=0$ so that the sum of rows/columns in $\mathbf{O}$ equals to zero, then we can easily obtain:
\begin{equation}
\begin{small}
\left\{
\begin{array}{lr}
\sum_{i=1}^{m}d_{ij}^{2} = \mathrm{Tr}(\mathbf{B}) + mb_{jj} &  \\
\sum_{j=1}^{m}d_{ij}^{2} = \mathrm{Tr}(\mathbf{B}) + mb_{ii} & \\
\sum_{i=1}^{m}\sum_{j=1}^{m}d_{ij}^{2} = 2m\mathrm{Tr}(\mathbf{B}) &  
\end{array},
\right.
\end{small}
\end{equation}
where $\mathrm{Tr}(\cdot)$ is the trace of matrix, $\mathrm{Tr}(B) = \sum_{i=1}^{m}\left \| o_i \right \|^2$. We denote:
\begin{equation}
\begin{small}
\left\{
\begin{array}{lr}
d_{i\cdot }^{2} = \frac{1}{m}\sum_{j=1}^{m}d_{ij}^2 &  \\
d_{\cdot j}^{2} = \frac{1}{m}\sum_{i=1}^{m}d_{ij}^2 & \\
d_{\cdot \cdot} = \frac{1}{m^2}\sum_{i=1}^{m}\sum_{j=1}^{m}d_{ij}^2 &  
\end{array},
\right.
\end{small}
\end{equation}
From Eqs. (2)$\sim$(4), we can obtain the inner product matrix $\mathbf{B}$ by the distance matrix $\mathbf{D}$ as:
\begin{equation}
\begin{small}
b_{ij} = -\frac{1}{2}\left ( d_{ij}^2 - d_{i\cdot }^2 - d_{\cdot j}^2 + d_{\cdot \cdot }^2 \right )\,.
\end{small}
\end{equation}

By applying eigenvalue decomposition (EVD) \cite{chonavel2003fast} with $\mathbf{B}$, we can easily obtain the $(n+k)$-dimensional embedded representation $\mathbf{O}$ that models the $(n+k)$-dimensional embedded manifold ($\mathbb{R}^{n+k}$).

\subsubsection{Manifold Structure Alignment}

With this obtained $\mathbf{O}$, we consider the alignment of manifold structures from semantic to visual FS. Specifically, we measure the similarity of the combined semantic feature representation $S^{p+e}$ (pre-defined $S^p$ combined with expanded $S^e$) and the embedded representation $\mathbf{O}$ by cosine distance. In order to achieve the alignment jointly with the semantic feature expansion, we build an regularization term to further guide the autoencoder based network as:
\begin{equation}
\begin{small}
\mathcal{L}_a = \sum_{i=1}^{l} \sum_{j=1}^{m} {\bf 1}\left [ y_{i} = c_{j} \right ] \cdot  \left [ 1 - \frac{S_{i}^{p+e}\cdot o_{j}}{\left \| S_{i}^{p+e} \right \|\left \| o_{j} \right \|}  \right ]\,,
\end{small}
\end{equation}
where $S_{i}^{p+e}$ is the combined semantic feature representation of the $i$-th seen class example $x_{i}$, $y_{i}$ is the class label and $c_{j}$ is the $j$-th class label among $m$ classes. ${\bf 1}\left [ y_{i} = c_{j} \right ]$ is an indicator function that takes a value of one if its argument is true and zero otherwise. Lastly, combine with Eq. (1), the unified objective function can be described as:
\begin{equation}
\begin{tiny}
\mathcal{L} = \alpha \cdot \underset{\mathcal{L}_r}{\underbrace{\sum_{i=1}^{l}\left \| x_{i} - \hat{x_{i}} \right \|_{2}^{2}}} + \beta \cdot \underset{\mathcal{L}_a}{\underbrace{\sum_{i=1}^{l} \sum_{j=1}^{m} {\bf 1}\left [ y_{i} = c_{j} \right ] \cdot  \left [ 1 - \frac{S_{i}^{p+e}\cdot o_{j}}{\left \| S_{i}^{p+e} \right \|\left \| o_{j} \right \|}  \right ]}}\,,
\end{tiny}
\end{equation}
where $\mathcal{L}_r$ acts as a base term that mainly guides to reconstruct the visual input examples. $\mathcal{L}_a$ is an adaptation term that mainly guides the learning of latent vectors and forces the manifold structure of the combined semantic FS to approximate with the structure of embedded manifold extracted from visual FS. The $\alpha$ and $\beta$ are two hyper parameters that control the balance between them.

\begin{table*}[t]
    \renewcommand\thetable{2}
    \centering
    \fontsize{8.5}{0.1}\selectfont 
    \begin{threeparttable}  
        \caption{Comparison with state-of-the-art competitors}  
        \label{table2}  
        \begin{tabular}{lcccccccccc}  
            \toprule  
            \multirow{2}{*}{Method}&  
            \multicolumn{2}{c}{AWA}&\multicolumn{2}{c}{CUB}&\multicolumn{2}{c}{aPa\&Y}&\multicolumn{2}{c}{SUN}&\multicolumn{2}{c}{ImageNet}\cr  
            \cmidrule(lr){2-3} \cmidrule(lr){4-5} \cmidrule(lr){6-7} \cmidrule(lr){8-9} \cmidrule(lr){10-11} 
            &SS  &ACC               &SS   &ACC              &SS   &ACC                 &SS   &ACC              &SS   &ACC\cr  
            \midrule  
            DeViSE \cite{frome2013devise} ($'13$)           &A/W &56.7/50.4         &A/W  &33.5             &-    &-                   &-    &-                &A/W  &12.8\cr
            DAP \cite{lampert2014attribute} ($'14$)              &A   &60.1              &A    &-                &A    &38.2                &A    &72.0             &-    &-\cr
            MTMDL \cite{yang2014unified} ($'14$)            &A/W &63.7/55.3         &A/W  &32.3             &-    &-                   &-    &-                &-    &-\cr
            ESZSL \cite{romera2015embarrassingly} ($'15$)            &A   &75.3              &A    &48.7             &A    &24.3                &A    &82.1             &-    &-\cr
            SSE \cite{zhang2015zero} ($'15$)              &A   &76.3              &A    &30.4             &A    &46.2                &A    &82.5             &-    &-\cr
            RRZSL \cite{shigeto2015ridge} ($'15$)            &A   &80.4              &A    &52.4             &A    &48.8                &A    &84.5             &W    &-\cr
            Ba et al. \cite{ba2015predicting} ($'15$)        &A/W &69.3/58.7         &A/W  &34.0             &-    &-                   &-    &-                &-    &-\cr    
            AMP \cite{bucher2016improving} ($'16$)              &A+W &66.0              &A+W  &-                &-    &-                   &-    &-                &A+W  &13.1\cr
            JLSE \cite{zhang2016zero} ($'16$)             &A   &80.5              &A    &41.8             &A    &50.4                &A    &83.8             &-    &-\cr
            SynC$^{struct}$ \cite{changpinyo2016synthesized} ($'16$)  &A   &72.9              &A    &54.4             &-    &-                   &-    &-                &-    &-\cr
            MLZSC \cite{bucher2016improving} ($'16$)            &A   &77.3              &A    &43.3             &-    &53.2                &A    &84.4             &-    &-\cr
            SS-voc \cite{fu2016semi} ($'16$)           &A/W &78.3/68.9         &A/W  &-                &-    &-                   &-    &-                &A/W  &16.8\cr
            SAE \cite{kodirov2017semantic} ($'17$)              &A   &84.7              &A    &61.2             &A    &55.1                &A    &91.0             &W    &26.3\cr
            CLN+KRR \cite{long2017zero} ($'17$)          &A   &81.0              &A    &58.6             &-    &-                   &-    &-                &-    &-\cr
            MFMR \cite{xu2017matrix} ($'17$)             &A   &76.6              &A    &46.2             &A    &46.4                &A    &81.5             &-    &-\cr
            RELATION NET \cite{yang2018learning} ($'18$)     &A   &84.5              &A    &62.0             &-    &-                   &-    &-                &-    &-\cr
            CAPD-ZSL \cite{rahman2018unified} ($'18$)         &A   &80.8              &A    &45.3             &A    &55.0                &A    &87.0             &W    &23.6\cr
            LSE \cite{yu2018zero} ($'18$)              &A   &81.6              &A    &53.2             &A    &53.9                &-    &-                &W    &{\bf27.4}\cr
            AMS-SFE (Ours)           &A   &{\bf 90.9}        &A    &{\bf 67.8}       &A    &{\bf 59.4}          &A    &{\bf 92.7}       &W    &26.1\cr  
            \bottomrule  
        \end{tabular}
        \footnotesize{SS is Semantic Space, A is Attribute and W is Word Vectors; '/' means 'or' and '+' means 'and'; '-' means that there is no reported result. ACC is accuracy (\%) where Hit@1 is used for AWA, CUB, aPa\&Y, and SUN, Hit@5 is used for ImageNet}
    \end{threeparttable}
    \vspace{-0.2cm}  
\end{table*}  

\subsubsection{Prototype Update}

To update the prototypes, we have different strategies regarding to seen and unseen classes.

\noindent{\bf Seen Class Prototypes:} Because we have already obtained the trained autoencoder by optimizing Eq. (7), so we compute the center (i.e., the mean value) of all latent vectors $z_{i}$ belonging to the same class as $P^{e} = \frac{1}{h}\sum_{i=1}^{h}z_{i}$, and combine with the pre-defined prototype to update the prototype for each seen class as $P = P^{p} \uplus P^{e}$. Where $z_{i}$ is the expanded semantic features obtained by $Encoder(x_{i})$, $h$ is the number of examples belonging to one specific seen class, $P^{p}$ and $P^{e}$ are pre-defined and expanded semantic prototype for one specific class, and $\uplus$ concatenates/combines two vectors.

\noindent{\bf Unseen Class Prototypes:} As no unseen class example is available during training, so we cannot apply the $Encoder(\cdot)$ to expand the semantic features directly. Instead, we consider another strategy by utilizing the local linearity among prototypes. Specifically, for each pre-defined unseen class prototype, we first obtain its $g$ nearest neighbors from pre-defined seen class prototypes. Then we estimate this pre-defined unseen class prototype by a linear combination of these $g$ neighbors as:
\begin{equation}
\begin{small}
\begin{aligned}
{P^{p}}' &= \theta _{1}P_{1}^{p} + \theta _{2}P_{2}^{p} + \cdots +\theta _{g}P_{g}^{p} = \theta P_{1\rightarrow g}^{p}\,,  
\end{aligned}
\end{small}
\end{equation}
where ${P^{p}}'$ is the pre-defined prototype of one specific unseen class, $\left \{ P_{i}^{p} \right \}_{i=1}^{g}$ are its $g$ nearest neighbors from pre-defined seen class prototypes and $\left \{ \theta_{i} \right \}_{i=1}^{g}$ are the estimation parameters. This is a simple linear programming and can be solved easily by:
\begin{equation}
\begin{small}
\theta = \underset{\theta}{\arg \min} \left \| {P^{p}}'- \theta P_{1\rightarrow g}^{p} \right \|\,.
\end{small}
\end{equation}
With the obtained $\theta$, we update the class prototype for this unseen class as:
\begin{equation}
\begin{small}
\begin{aligned}
{P^{e}}' &= \theta _{1}P_{1}^{e} + \theta _{2}P_{2}^{e} + \cdots +\theta _{g}P_{g}^{e} \\
&= \theta P_{1\rightarrow g}^{e}\,,
\end{aligned} 
\end{small}
\end{equation}
\begin{equation}
\begin{small}
{P}' = {P^{p}}' \uplus  {P^{e}}'\,,
\end{small}
\end{equation}
where ${P}'$ is the updated prototype for this unseen class and $\left \{ P_{i}^{e} \right \}_{i=1}^{g}$ are the corresponding $g$ neighbor expanded seen class prototypes.

\subsubsection{Testing Recognition}
In our model, similar to some methods, we also adopt the simple semantic autoencoder training framework \cite{kodirov2017semantic} to learn the projection between the visual and semantic FS. As to the recognition for unseen class example, we simply search the most closely related prototype with its projected semantic features, and set the class corresponding with this prototype to the unseen class example. The recognition is described as:
\begin{equation}
\begin{small}
\Omega  ({x_{i}}') = \underset{j}{\arg \min}\,Dist(f_{e}({x_{i}}'), {P_{j}}')\,,
\end{small}
\end{equation}
where ${x_{i}}'$ is the testing unseen class example, $f_{e}(\cdot)$ is the trained projection that projects ${x_{i}}'$ to the semantic FS, ${P_{j}}'$ is the prototype for the $j$-th unseen class, $Dist(\cdot, \cdot)$ is a distance measurement and $\Omega(\cdot)$ returns the class label.
%

\section{Experiment}

\subsection{Settings}

\noindent{\bf Datasets:} Our model is evaluated on five widely used benchmark datasets for ZSL including Animals with Attributes (AWA) \cite{lampert2014attribute}, CUB-200-2011 Birds (CUB) \cite{wah2011caltech}, aPascal\&Yahoo (aPa\&Y) \cite{farhadi2009describing}, SUN Attribute (SUN) \cite{patterson2014sun} and ILSVRC2012/ILSVRC2010 (ImageNet) \cite{russakovsky2015imagenet}. The basic description of them is listed in Table 1.

\begin{table}[h]
\vspace{-0.5cm}
\renewcommand\thetable{1}
    \begin{center}
        \caption{Description of datasets. Notation: \# -- number, SCs/UCs -- seen/unseen classes, D-SF -- dimension of semantic feature.}
        \setlength{\tabcolsep}{2.7mm}{  
            \begin{tabular}{ccccc}        
                \hline                   
                Dataset         & \# Examples       & \# SCs          & \# UCs            & D-SF \\
                \hline
                AWA             & 30475             & 40              & 10                & 85    \\ 
                CUB             & 11788             & 150             & 50                & 312   \\ 
                aPa\&Y          & 15339             & 20              & 12                & 64    \\
                SUN             & 14340             & 645             & 72                & 102    \\
                ImageNet        & $2.54\times 10^5$ & 1000            & 360               & 1000  \\
                \hline  
        \end{tabular}}
    \end{center}  
\vspace{-0.5cm}
\end{table}

\noindent{\bf Competitors:} We compare our model with 18 state-of-the-art competitors (Table 2). These methods are all proposed most recently and cover a wide range of models. All methods are under the same settings on datasets, evaluation criterion and non-transductive setting.

\noindent{\bf Evaluation Criterion:} As common practice in ZSL, we use Hit@k accuracy \cite{frome2013devise} to evaluate models. The model predicts top-k possible class labels of one testing unseen class example, and it correctly classifies the example if and only if the ground truth is within these k class labels. Hit@1 is evaluated for AWA, CUB, aPa\&Y, and SUN, which is the ordinary accuracy, and Hit@5 is evaluate for ImageNet.

\noindent{\bf Implementation:} In our experiment, the features we use are extracted from GoogleNet \cite{szegedy2015going} for the visual FS. Each image example is presented by a 1024-dimensional vector. As to the semantic FS, semantic attributes are used for AWA, CUB, aPa\&Y, and SUN, and semantic word vectors are used for ImageNet. The autoencoder based network for expansion and alignment is with five hidden layers, and one input/output layer respectively. The central hidden layer is adjusted to the dimension of semantic features we expand. 65, 138, 26, 58, 12 for AWA, CUB, aPa\&Y, SUN, and ImageNet, respectively. As to hyper parameters $\alpha$ and $\beta$, we choose 9 and 77 respectively by grid-search.

\noindent{\bf Non-Transductive:} As mentioned in Section 2, our model strictly complies with the zero-shot setting that the training only relies on seen class examples, and the unseen class examples are solely available during testing phase.

\subsection{Results and analysis}
\noindent{\bf General Results:} The comparison results with these state-of-the-art competitors are shown in Table 2. Our model outperforms all competitors with great advantages in AWA, CUB, aPa\&Y, and SUN. The accuracy achieves 90.9\%, 67.8\%, 59.4\% and 92.7\%, respectively. While in ImageNet, due to the expandable auxiliary features are limited, our model is slightly weaker (-1.3\%) than the strongest competitor. From Tabel 1, we can observe the D-SF for ImageNet is 1000, while the dimension of its visual feature is 1024. So in our model, the expandable auxiliary features for ImageNet ($\left [ 0, 24 \right ]$) are far less than the pre-defined ones, which makes the difficulty of alignment. Instead, we have enough expandable auxiliary features for the other four datasets, so the alignment can be better approximated. The dimension of pre-defined and expanded features for 5 datasets is shown in Table 3.
\begin{table}[h]
\vspace{-0.4cm}
\renewcommand\thetable{3}
    \begin{center}
        \caption{Dimension of pre-defined (P) / expanded (E) features}
        \setlength{\tabcolsep}{1.8mm}{  
            \begin{tabular}{cccccc}        
                \hline                   
                           & AWA     & CUB     & aPa\&Y  & SUN     & ImageNet\\
                \hline
                P                & 85      & 312     & 64      & 102     & 1000 \\
                E                & 65      & 138     & 26      & 58      & 12 \\
                P+E              & 150     & 450     & 90      & 160     & 1012 \\ 
                \hline  
        \end{tabular}}
    \end{center}  
\vspace{-0.5cm}
\end{table}

\noindent{\bf Projection Robustness:} We conduct the evaluation on AWA and compare with the strongest competitor SAE \cite{kodirov2017semantic} to verify the projection robustness of our model. A projection that maps from the visual to semantic FS is trained on seen class examples with our model. Then we apply this projection to all testing unseen class examples and obtain their semantic feature representations. We visualize the obtained semantic feature representations by t-SNE \cite{maaten2008visualizing}, and the results are shown in Fig.2 and Fig.3. The former is the result of SAE and the latter is the result of our model. In our model, only a small percentage of these testing unseen class examples are mis-projected. And due to the implicit alignment of manifold structures from semantic to visual FSs, these mis-projected examples are less shifted. This means that our model can obtain better results for Hit@k accuracy when k varies, as shown in Table 4. 

\begin{table}[h]
\vspace{-0.4cm}
\renewcommand\thetable{4}
    \begin{center}
        \caption{Hit@k accuracy (\%) for AWA, $k\in \left [ 1,5 \right ]$}
        \setlength{\tabcolsep}{0.9mm}{  
            \begin{tabular}{cccccc}        
                \hline                   
                Method           & Hit@1   & Hit@2   & Hit@3   & Hit@4   & Hit@5\\
                \hline
                SAE              & 84.7    & 93.5    & 97.2    & 98.8    & 99.4 \\
                AMS-SFE (ours)   & 90.9    & 97.4    & 99.5    & 99.8    & 99.8 \\ 
                \hline  
        \end{tabular}}
    \end{center} 
\vspace{-0.5cm} 
\end{table}

\begin{table}[h]
\vspace{-0.5cm}
\renewcommand\thetable{5}
    \begin{center}
        \caption{Ablation comparison (accuracy\%) on pre-defined (P) / expanded (E) semantic features and Both (P+E)}
        \setlength{\tabcolsep}{1.9mm}{  
            \begin{tabular}{cccccc}        
                \hline                   
                           & AWA     & CUB     & aPa\&Y  & SUN     & ImageNet\\
                \hline
                P                & 84.4    & 60.3    & 53.1    & 88.7.0  & 26.1 \\
                E                & 75.2    & 52.8    & 45.5    & 77.4    & 14.2 \\
                P+E              & 90.9    & 67.8    & 59.4    & 92.7    & 26.1 \\ 
                \hline  
        \end{tabular}}
    \end{center} 
\vspace{-0.5cm} 
\end{table}

\noindent{\bf Ablation Comparison:} To further evaluate the effectiveness of our model, we conduct the ablation experiment. We compare the performance on five benchmark datasets on three scenarios as follows. (1) Only pre-defined semantic features are used; (2) Only expanded semantic features are used; (3) Both of them are used and under alignment. The results are shown in Table 5, our model greatly improves the performance of ZSL by doing alignment with the expanded semantic features.

\noindent{\bf Fine-Grained Accuracy:} We record and count the prediction for each testing unseen class example. We also conduct the evaluation on AWA and compare with the strongest competitor SAE \cite{kodirov2017semantic}. The results are presented by confusion matrix (CM), where Fig.4 and Fig.5 show the confusion matrix of SAE and our model respectively. In the confusion matrix, the diagonal position indicates the classification accuracy for each class, the column means the ground truth and the row denotes the predicted results. It can be seen that our model obtains higher accuracy, along with more balanced and robust prediction results for each testing unseen class.

\begin{figure}[t]
    \centering  
    \begin{minipage}[t]{0.49\linewidth}
        \renewcommand\thefigure{2}  
        \centering   
        \includegraphics[width=1.6in]{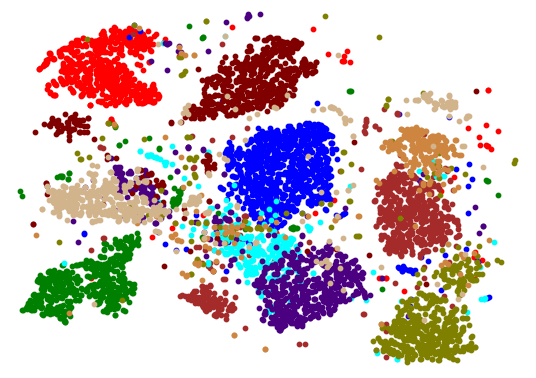}
        \caption{Projection-SAE}  
        \label{fig5}  
    \end{minipage}
    \begin{minipage}[t]{0.49\linewidth}
        \renewcommand\thefigure{3}  
        \centering   
        \includegraphics[width=1.6in]{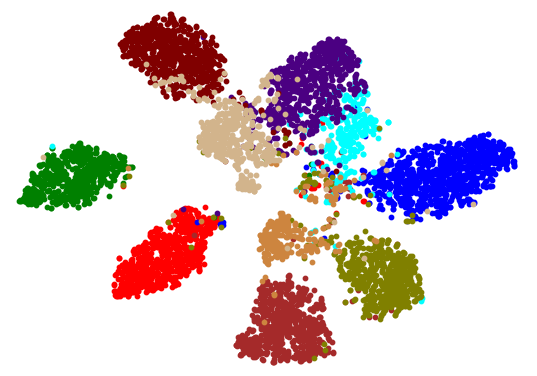}  
        \caption{Projection-AMS-SFE}  
        \label{fig5}  
    \end{minipage}    
\end{figure}

\begin{figure}[t]
    \centering  
    \begin{minipage}[t]{0.49\linewidth}
        \renewcommand\thefigure{4}  
        \centering   
        \includegraphics[width=1.6in]{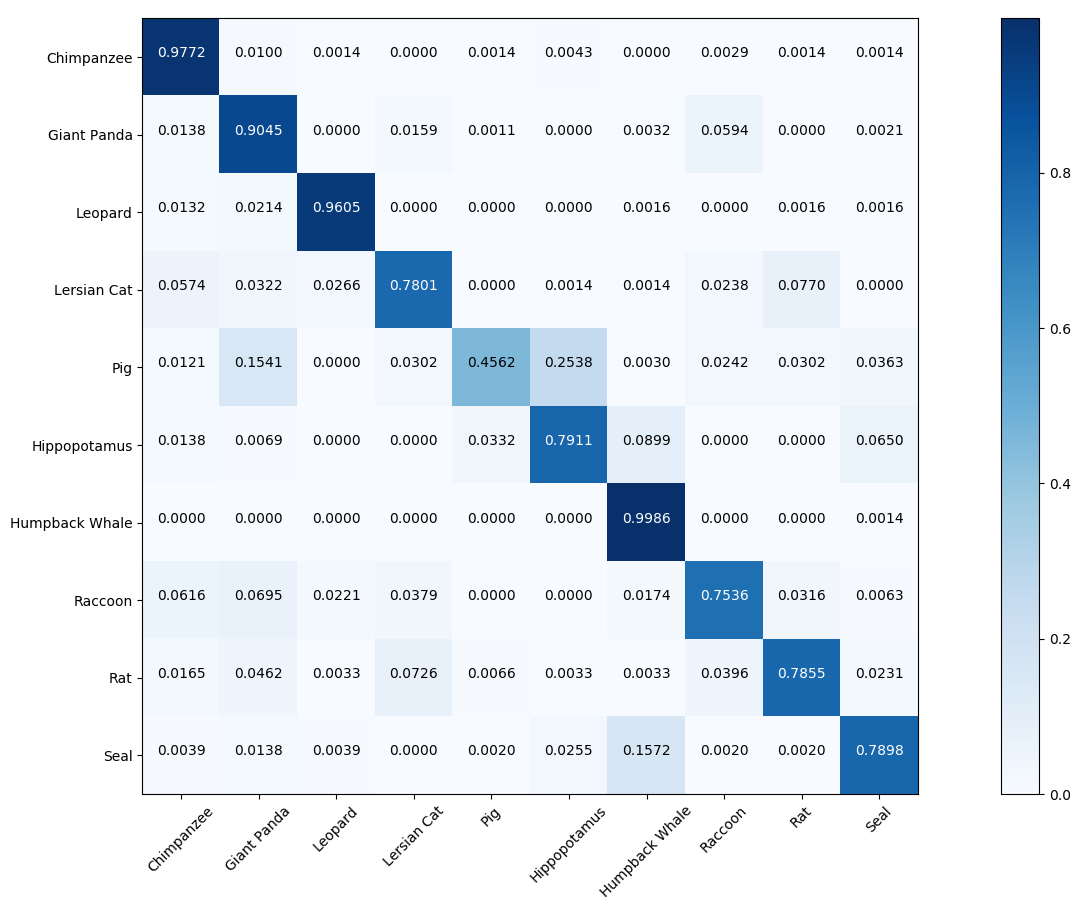}
        \caption{CM-SAE}  
        \label{fig5}  
    \end{minipage}
    \begin{minipage}[t]{0.49\linewidth}
        \renewcommand\thefigure{5}  
        \centering   
        \includegraphics[width=1.6in]{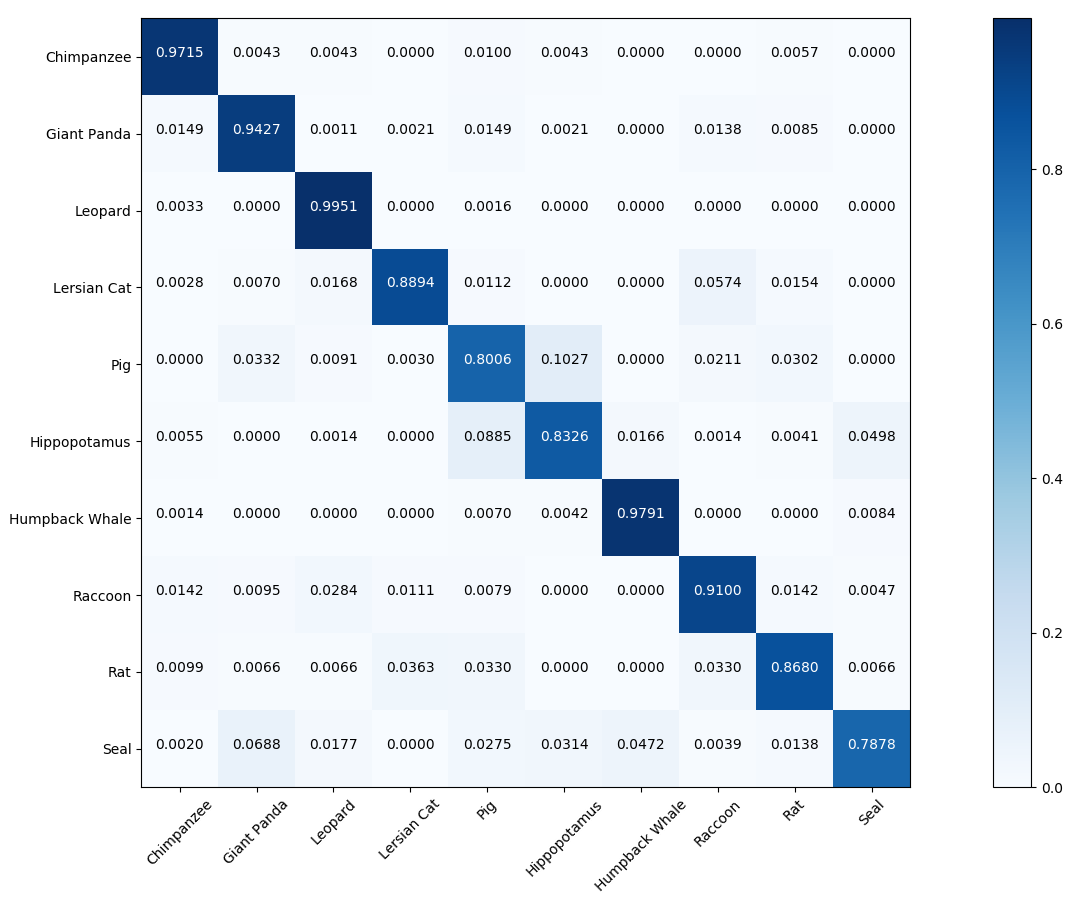}  
        \caption{CM-AMS-SFE}  
        \label{fig5}  
    \end{minipage}
\vspace{-0.5cm}      
\end{figure}

\section{Conclusion}
In this paper, we proposed a novel model (AMS-SFE) for zero-shot learning that considers aligning the manifold structures of the semantic and visual feature spaces by jointly conducting semantic feature expansion. Our model can better mitigate the domain shift problem and obtain a more robust and generalized projection between the visual and semantic feature spaces. In the future, we plan to investigate the more efficient and generalized way to further empower the semantic feature space in zero-shot learning.

{\small
\bibliographystyle{IEEEbib}
\bibliography{my}
}
\end{document}